\documentclass[letterpaper]{article} 
\usepackage{aaai2026}  
\usepackage{times}  
\usepackage{helvet}  
\usepackage{courier}  
\usepackage[hyphens]{url}  
\usepackage{graphicx} 
\usepackage{natbib}  
\usepackage{caption} 
\usepackage{algorithm}
\usepackage{algorithmic}

\usepackage{amsmath}
\usepackage{amssymb}
\usepackage{bigstrut}
\usepackage{booktabs}
\usepackage{multirow}
\usepackage{makecell}
\usepackage{subcaption}

\usepackage{newfloat}
\usepackage{listings}
\DeclareCaptionStyle{ruled}{labelfont=normalfont,labelsep=colon,strut=off} 
\floatstyle{ruled}
\newfloat{listing}{tb}{lst}{}
\floatname{listing}{Listing}

\title{SC-Net: Robust Correspondence Learning via Spatial and Cross-Channel Context}
\author{
    Shuyuan Lin\textsuperscript{\rm 1}, 
    Hailiang Liao\textsuperscript{\rm 1}, 
    Qiang Qi\textsuperscript{\rm 2}, 
    Junjie Huang\textsuperscript{\rm 1}, 
    Taotao Lai\textsuperscript{\rm 3}, 
    Jian Weng\textsuperscript{\rm 1}\thanks{Corresponding Author.} 
}
\affiliations{
    \textsuperscript{\rm 1}College of Cyber Security, Jinan University, Guangzhou, China\\
    \textsuperscript{\rm 2}School of Data Science, Qingdao University of Science and Technology, Qingdao, China\\
    \textsuperscript{\rm 3}School of Computer and Data Science, Minjiang University, Fuzhou, China\\
    swin.shuyuan.lin@gmail.com
}

\usepackage{bibentry}

\begin{document}

\maketitle

\begin{abstract}
Recent research has focused on using convolutional neural networks (CNNs) as the backbones in two-view correspondence learning, demonstrating significant superiority over methods based on multilayer perceptrons. However, CNN backbones that are not tailored to specific tasks may fail to effectively aggregate global context and oversmooth dense motion fields in scenes with large disparity. To address these problems, we propose a novel network named SC-Net, which effectively integrates bilateral context from both spatial and channel perspectives. Specifically, we design an adaptive focused regularization module (AFR) to enhance the model's position-awareness and robustness against spurious motion samples, thereby facilitating the generation of a more accurate motion field. We then propose a bilateral field adjustment module (BFA) to refine the motion field by simultaneously modeling long-range relationships and facilitating interaction across spatial and channel dimensions. Finally, we recover the motion vectors from the refined field using a position-aware recovery module (PAR) that ensures consistency and precision. Extensive experiments demonstrate that SC-Net outperforms state-of-the-art methods in relative pose estimation and outlier removal tasks on YFCC100M and SUN3D datasets. 
\end{abstract}


\begin{links}
\link{Code}{http://www.linshuyuan.com}
\end{links}

\section{Introduction}
\begin{figure*}
	\centering
    \includegraphics[width=\textwidth]{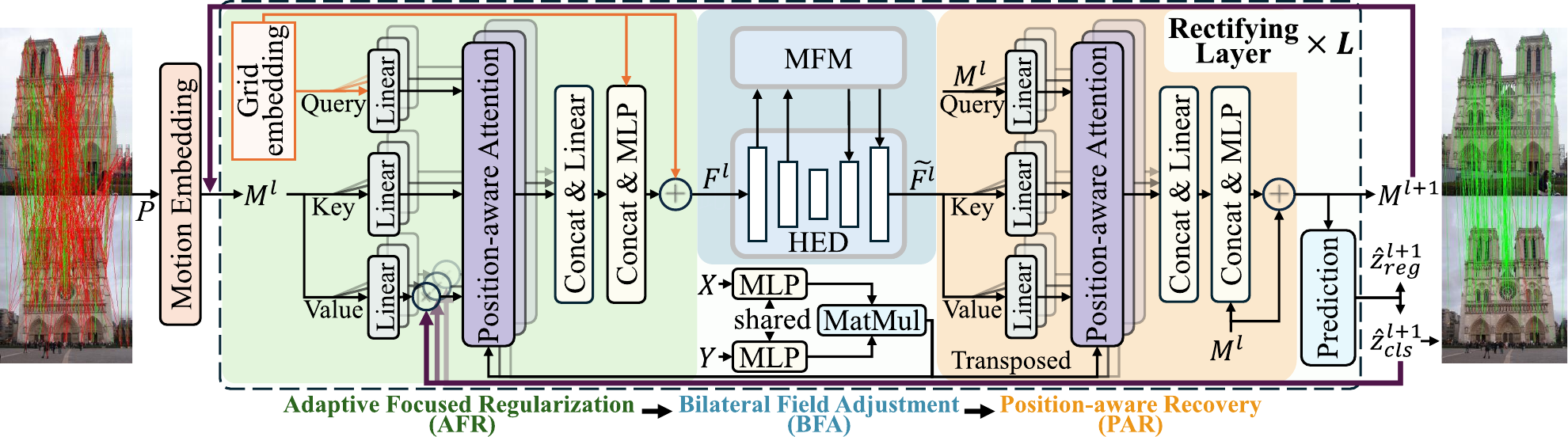}
	\caption{Architecture of the proposed SC-Net. It consists of $L$ rectifying layers, each including AFR, BFA, and PAR. SC-Net takes putative correspondences $P\in\mathbb{R}^{N\times 4}$ as input and predicts the inlier probabilities $\hat{z}_{cls}^{l+1}$ and regression weights $\hat{z}_{reg}^{l+1}$. Purple lines denote propagation to subsequent layer.}
	\label{fig:architecture}
\end{figure*}

Two-view correspondence, a core task in computer vision, aims to establish reliable matches between image pairs for recovering camera geometry. It underpins many downstream applications such as panoramic stitching~\cite{brown2007automatic}, simultaneous localization and mapping~\cite{placed2023survey} and structure-from-motion~\cite{schonberger2016structure}. However, due to the limited discriminative power of local descriptors~\cite{lin2024robust}, putative matches often contain outliers, especially under repetitive patterns~\cite{mousavi2022two}, large viewpoint changes~\cite{jin2021image}, motion blur and occlusion~\cite{lin2024multimotion}.

Recently, ConvMatch~\cite{zhang2023convmatchaaai} introduced a convolutional neural network (CNN)-based framework with stacked ResBlocks~\cite{he2016resnet} to extract deep features for correspondence learning, replacing earlier multilayer perceptron (MLP)-based designs. 
While effective, it struggles to capture long-range dependencies and global context due to the limited receptive field of convolutions~\cite{zhao2021clnet,li2025loki}. 
ConvMatch\textsuperscript{+}~\cite{zhang2023convmatchtpami} improves upon this by using bilateral convolutions and SE modules~\cite{hu2018squeeze} to preserve channel-level information. However, it still lacks spatially-aware global reasoning and remains constrained by local convolutional operations.
Both versions employ a graph attention network (GAT)~\cite{velivckovic2017graph} for structuring motion vectors, but repeated message passing leads to fine-grained spatial information loss and further exacerbates over-smoothing~\cite{chen2020measuring}.

To address the challenges, we propose a new framework, \textbf{SC-Net}, which partitions space into grids to explicitly encode spatial priors, and utilizes classification priors and shared positional bias to prevent the loss of spatial information in deeper GATs (as discussed in~\cite{lindenberger2023lightglue}) — an issue not addressed by previous methods. Specifically, SC-Net mainly benefits from two key modules.
The first is the \textbf{Adaptive Focus Regularization module (AFR)}, which facilitates sharper and cleaner message passing along all paths from sample to field.
The second is the \textbf{Bilateral Field Adjustment module (BFA)} , which establishes bilateral global dependencies across both spatial and channel dimensions, thereby enhancing information interaction between local motion regions and dynamically refining their central motion features. 
Finally, to ensure consistency and precision, the position-aware recovery module (PAR) is applied as a standard refinement step to recover the final motion vectors from the smoothed field.
Collectively, these modules enable each location in the motion field to move beyond local receptive field constraints, attend to relevant positions and better preserve discontinuities across motion patterns at varying scales~\cite{zhang2024dematch}.
As shown in Fig.~\ref{fig:architecture}, our main contributions are summarized as follows:
\begin{itemize}
	\item We present SC-Net, a novel network that integrates position-aware attention and cross-channel context modeling to preserve fine-grained local structures for correspondence pruning. In contrast to methods like OANet, SC-Net explicitly encodes spatial priors and achieves superior performance across diverse scenarios.
    \item We propose AFR to obtain a clean initial motion field from contaminated motion samples, simplifying the complexity of subsequent rectification while overcoming the over-smoothing issue encountered in GNNs.
    \item We propose BFA, which models long-range relationships across spatial and channel dimensions, enhancing the network’s matching capability in complex scenarios.
\end{itemize}

\section{Related Work}
\subsection{Traditional Outlier Removal Methods}
Traditional outlier removal approaches can be broadly categorized into three classes: resampling-based, non-parametric, and relaxed methods. Among them, resampling methods like RANSAC~\cite{fischler1981random} and its variants~\cite{barath2018gcransac,barath2020magsac++} follow a hypothesize-and-verify paradigm for robust model estimation. While effective under moderate noise, they struggle with complex transformations and heavily contaminated data.
To overcome the rigidity of parametric models, non-parametric methods such as VFC~\cite{ma2014robust} and SparseVFC~\cite{ma2013regularized} learn a smooth vector field under generalized geometric constraints, with SparseVFC introducing sparse approximations for improved efficiency. However, their reliance on smoothness priors limits performance in scenes with large motion discontinuities.
Relaxed methods aim to handle more diverse scenarios by loosening geometric constraints. For instance, CODE~\cite{lin2017code} integrates a coherence-aware likelihood model, while LPM~\cite{ma2019locality} and GMS~\cite{bian2017gms} exploit local consistency among matches.
Despite their respective strengths, these methods require careful parameter tuning and often fail in the presence of strong outliers or occlusions.

\subsection{Learning-based Outlier Removal Methods}
Learning-based outlier rejection has advanced rapidly with the rise of deep networks~\cite{li2024learning,lin2025mgcanet}.
As an early attempt, LFGC~\cite{yi2018lfgc} formulates outlier rejection as a binary classification task using MLPs, but struggles to capture local context.
Subsequent methods enhance contextual modeling within MLP-based frameworks. OANet~\cite{zhang2019oanet} incorporates spatial correlation and pooling to model local-global dependencies. 
T-Net~\cite{zhong2021tnet} and NCMNet~\cite{liu2023ncmnet} further improve context encoding via channel-wise attention and neighbor space interaction. 
MSGSA~\cite{lin2024msgsa} extends this by leveraging inter-stage consistency.
To break from the limitations of MLPs, ConvMatch~\cite{zhang2023convmatchaaai,zhang2023convmatchtpami} introduces a CNN backbone for better geometric perception. PT-Net~\cite{gong2024ptnet} goes further by stacking CNNs and Transformers in a pyramid to capture both local and global dependencies. DeMatch~\cite{zhang2024dematch} reformulates outlier rejection by decomposing disordered motions into dominant flows. While effective, existing methods still face challenges in modeling complex motion patterns and long-range dependencies. To address this, we propose a novel network that integrates spatial awareness and cross-channel context to improve both matching accuracy and motion field consistency.

\section{Methodology}
\subsection{Overview}
Given a pair of images $(I, I')$, keypoints and descriptors are first extracted by off-the-shelf methods such as SIFT~\cite{lowe2004distinctive}. A nearest neighbor matcher is then used to generate an initial correspondence set $P$ across images:
\begin{equation}
	P=\{(x_i,x'_i)  \mid i=1,2,\cdots,N\}\in \mathbb{R}^{N\times 4}, 
\end{equation}
where $N$ is the number of correspondences, and $(x_i,x'_i)$ indicates the $i$-th pair of normalized keypoint coordinates in $I$ and $I'$, respectively. In practice, $P$ typically contains a large number of outliers. Therefore, our goal is to accurately identify inliers and estimate the relative camera motion.

We then embed the motion displacement and corresponding coordinates in the high-dimensional feature space to obtain the initial unordered motion vectors:
\begin{equation}
	M^0=\{\mathcal{F}_1(d_i)+\mathcal{F}_2(x_i) | i=1,2,\cdots,N\}\in \mathbb{R}^{N\times C},
\end{equation} 
where $d_i=x'_i-x_i$ indicates the displacement between keypoint $x_i$ and $x'_i$; $\mathcal{F}_1(\cdot)$ and $\mathcal{F}_2(\cdot)$ denote different MLPs that progressively elevate the dimensionality to extract deep features; $C$ represents the dimension of motion vectors.

After that, the unordered motion vectors are fed into multiple consecutive rectifying layers to correct them and predict the inlier logits $\hat{z}_{cls}^{l}$ and regression weights $\hat{z}_{reg}^{l}$:
\begin{equation}
	M^l, \hat{z}_{cls}^{l}, \hat{z}_{reg}^{l} = h_{\theta_l}(M^{l-1}), \quad l = 1, 2, \cdots, L,
\end{equation}
where $h_{\theta_l}(\cdot)$ indicates the $l$-th rectifying layer with learnable parameters $\theta_l$, and $L$ denotes the total number of rectifying layers in the network. $\hat{z}_{cls}^{l}$ guides the next layer, whereas $\hat{z}_{cls}^{L}$ is used to classifies correspondences.
For the regression task, we calculate the confidence scores $\hat{c}$ based on $\hat{z}_{cls}^{L}$ and $\hat{z}_{reg}^{L}$ of the last layer and apply the weighted 8-point algorithm to estimate the parametric model~\cite{sun2020acne}.

\subsection{Adaptive Focused Regularization}
To achieve learnable regularization~\cite{zhang2023convmatchaaai}, we first define a bounded 2D space $\Omega=\{(u,v)\in \mathbb{R}^2|-1\leq u\leq 1,-1\leq v\leq 1\}$, where the coordinates of matching pairs are normalized based on the camera intrinsic parameters or image size. This space is uniformly partitioned into a $K \times K$ grid, whose cell centers (hereafter referred to as grid coordinates) are encoded via a MLP to obtain the grid embeddings $G = \{g_k \mid k = 1, 2, \ldots, K^2\} \in \mathbb{R}^{K^2 \times C}$.
We then construct a complete graph by connecting each grid embedding $g_k$ and all motion features $M^l$, and apply a graph attention network $\mathcal{G}$ to allow each cell to focus on motion samples relevant to its spatial region. This enables the model to capture region-specific motion patterns and perform initial motion estimation locally, as follows:
\begin{equation}
	F^{l}=\mathcal{G}(G,M^l),
	\label{eq:gat}
\end{equation}
where $G$ serves as the query input; $M^l$ acts as both the key and value inputs; $F^l$ denotes the estimated sparse motion field in the $l$-th rectifying layer.
It is worth noting that stacking additional rectifying layers degrades positional cues and increases sensitivity to outliers, resulting in spatial blurring and over-smoothing~\cite{chen2020measuring}.

To address these issues, we first employ a shared-weight MLP to embed the raw positions and compute positional correlations, which are used to adjust the attention score matrix and enhance the model’s spatial sensitivity. Additionally, we incorporate the classification logits $\hat{z}_{cls} \in \mathbb{R}^N$ from the preceding layer to weight the value embeddings, thereby mitigating the adverse effects of spurious motion samples. The attention paradigm in Eq.~(\ref{eq:gat}) can be replaced as:
\begin{equation}
	O_i=\operatorname{Softmax}(\frac{Q_iK_i^\mathsf{T}}{\sqrt{C_{qk}}}+\psi({\alpha_i}\cdot\frac{S}{\sqrt{C}}+\beta_i))\hat{Z}_{cls}V_i,
	\label{eq:afr}
\end{equation}
\begin{equation}
	S=\mathcal{F}_3(Y)\mathcal{F}_3(X)^\mathsf{T},
	\label{eq:afr2}
\end{equation}
\begin{equation}
	\hat{Z}_{cls}=\operatorname{Diag}(\sigma(\hat{z}_{cls})),
\end{equation}
where $Q_i$, $K_i$, $V_i$ denote the qeury, key and value matrices for the $i$-th head; $C_{qk}$ is the dimension of query and key; $\alpha_i$ and $\beta_i$ are the learnable scale factors and biases for $i$-th head, respectively; $S$ is the positional correlation matrix shared with all heads; $\psi(\cdot)$ is a leaky ReLU function; $X$ and $Y$ represent the motion coordinates and grid coordinates, respectively; $\sigma(\cdot)$ is a sigmoid activation. Overall, the proposed AFR utilizes two improvements, namely the position-aware attention and the soft filtering, which enable each local region to focus on high-confidence motion samples from nearby locations and adaptively aggregate beneficial context to construct a cleaner motion field.

\subsection{Bilateral Field Adjustment}

\begin{figure}
	\centering
	\includegraphics[width=1\linewidth]{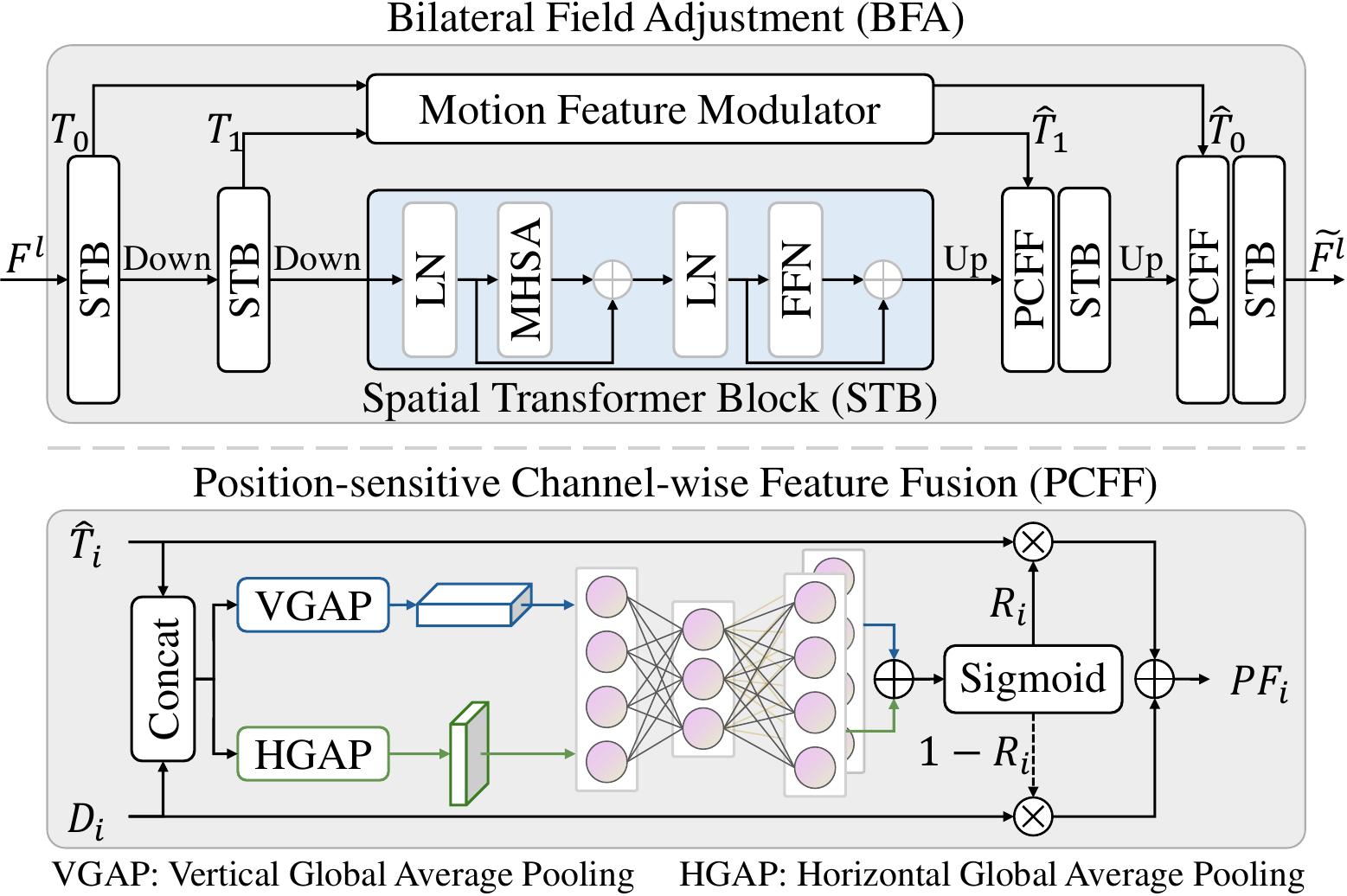}
	\caption{Structure of Bilateral Field Adjustment (BFA).}
	\label{fig:bfa}
\end{figure}

To avoid the over-smoothing effect of convolution-based rectification~\cite{zhang2023convmatchaaai,zhang2023convmatchtpami,gong2024ptnet}, we propose BFA, which combines a hierarchical encoder-decoder and a motion feature modulator to better preserve motion discontinuities, as illustrated in Fig.~\ref{fig:bfa}.

\noindent{\textbf{Hierarchical Encoder-Decoder.}} As shown in Fig.~\ref{fig:bfa} (a-b), the encoder-decoder primarily consists of spatial transformer blocks, with patch merging and patch expanding layers used for down-sampling and up-sampling~\cite{cao2022swinunet}, respectively. Based on these, we construct a hierarchical spatial encoder-decoder to simultaneously capture both low-level motion details and high-level motion pattern features, thereby enabling refined and piecewise smoothing across the motion field. To address the inconsistency between upsampled features and those from skip connections, we propose a novel Position-sensitive Channel-wise Feature Fusion (PCFF) module, which facilitates more accurate feature integration and improves transformer-based decoding. 
Specifically, PCFF first upsamples the extracted deep feature, denoted as $D_i$, and concatenates it with the corresponding skip-connected feature $\hat{T}_i$ to form $U_i = [D_i \| \hat{T}_i]$. Then, two 1D global average pooling operations~\cite{hou2021coordinate} are applied to extract direction-specific statistical cues, which are passed through two MLPs (denoted as $\mathcal{F}_h$ and $\mathcal{F}_w$) with shared nonlinear transformations. The resulting pixel-wise attention maps are used to recalibrate $\hat{T}_i$ and $D_i$, and to guide their weighted fusion, as follows:
\begin{equation}
\hat{U}_i^h=\mathcal{F}_h(\operatorname{Pool}_h(U_i)),\hat{U}_i^w=\mathcal{F}_w(\operatorname{Pool}_w(U_i)),
\end{equation}
\begin{equation}
	R_i=\sigma(\mathcal{S}(\hat{U}_i^h,\hat{U}_i^w)),
\end{equation}
\begin{equation}
	PF_i=R_i{\odot}\hat{T}_i+(\mathbf{1}-R_i){\odot}D_i,
\end{equation}
where $\operatorname{Pool}_h(\cdot)$ and $\operatorname{Pool}_w(\cdot)$ represent horizontal and vertical pooling operations, respectively; $\mathcal{S}(\cdot)$ denotes the broadcasting sum operation; $\odot$ indicates the Hadamard product.

\begin{figure}
	\centering
	\includegraphics[width=1\linewidth]{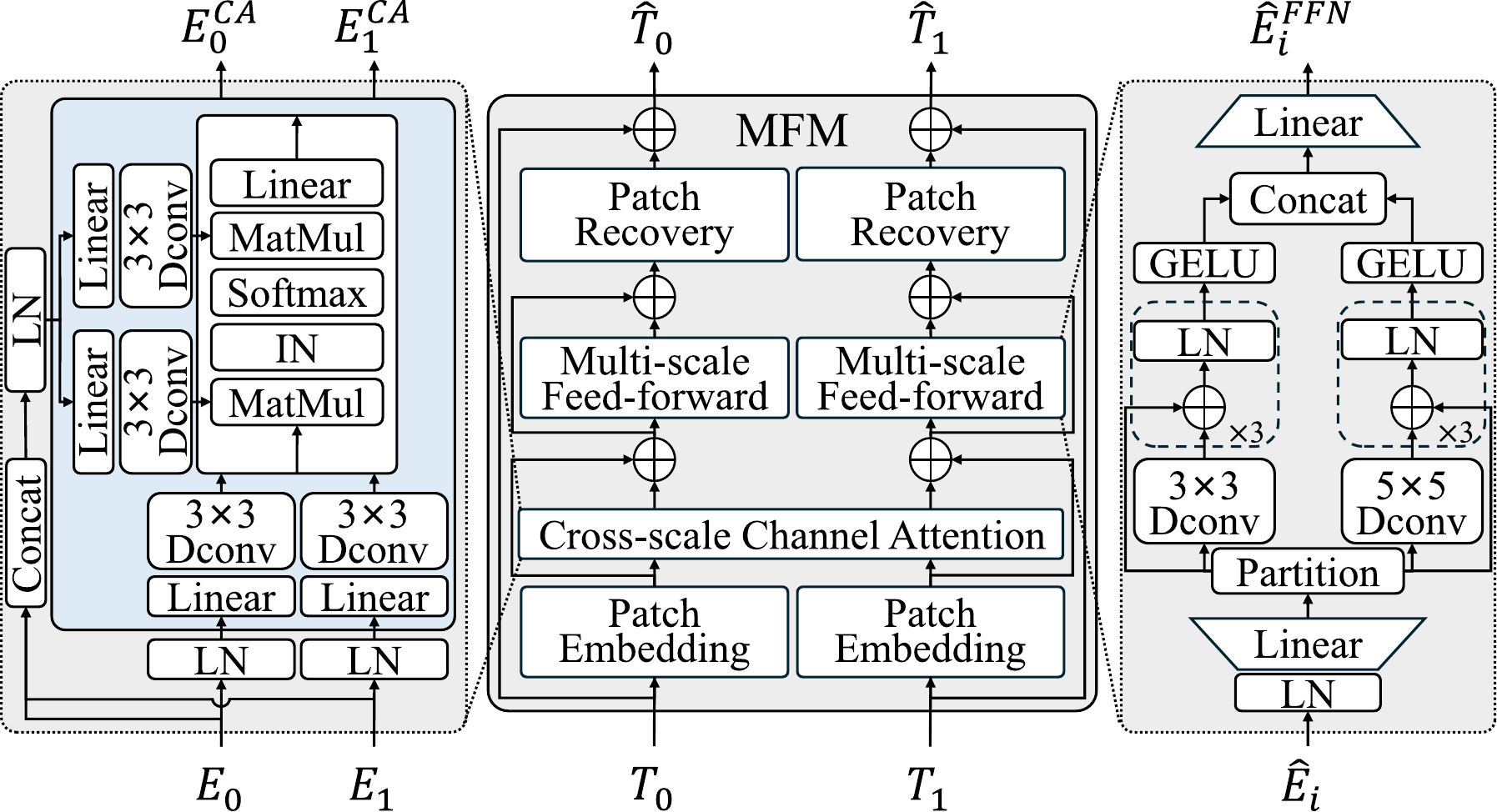}
	\caption{Structure of Motion Feature Modulator (MFM).}
	\label{fig:ms_seffn}
\end{figure}

\noindent{\textbf{Motion Feature Modulator.}} As shown in Fig.~\ref{fig:bfa}, the multi-scale encoded features $T_i$ $(i=0,1)$ are fed into the proposed Motion Feature Modulator (MFM), which modulates the channel-wise feature distribution to reduce the motion ambiguities across different scales.
To explain further (see Fig.~\ref{fig:ms_seffn}), MFM consists of two components: a cross-scale channel attention (CSCA) and a multi-scale feed-forward network (MSFFN). Given the multi-scale encoded features $T_i$, we first conduct spatial alignment by performing patch embedding on them, thereby obtaining flattened 2D patches $E_i{\in}\mathbb{R}^{P{\times}C_i}$. We then take these tokens as queries, treating their concatenation $E_{\sum}\in\mathbb{R}^{P\times\sum_iC_i}$ as keys and values. Similar to self-attention~\cite{vaswani2017attention}, we generate $Q_i^{CA}\in\mathbb{R}^{P\times C_i}$ and $\{K^{CA},V^{CA}\}\in\mathbb{R}^{P\times C_{\sum}}$ via layer normalization followed by linear projection. To facilitate the implicit modeling of the contextualized global relationships during the computation of covariance-based attention maps, we embed spatial context information into them using depth-wise convolution. This operation yields $\hat{Q}_i^{CA}\in\mathbb{R}^{P\times C_i}$ and $\{\hat{K}^{CA},\hat{V}^{CA}\}\in\mathbb{R}^{P\times C_{\sum}}$. 
Using these, we calculate the attention matrix to weight the values:
\begin{equation}
	E_i^{CA}=\operatorname{Linear}(\operatorname{Softmax}(\operatorname{IN}(\frac{(\hat{Q}_i^{CA})^\mathsf{T}\hat{K}^{CA}}{\sqrt{C_{\sum}}}))(\hat{V}^{CA})^\mathsf{T}),
\end{equation}
and apply a skip connection to obtain $\hat{E}_i=E_i+E_i^{CA}$, where $\operatorname{IN}(\cdot)$ denotes instance normalization used for steadily propagating the gradient during training~\cite{wang2022uctransnet}. Traditional feed-forward networks (FFNs) enhance representational capabilities by independently applying non-linear transformations to each pixel position, but they neglects the spatial dependencies between pixels. Therefore, the proposed MSFFN extends the original single path into a dual-path structure and integrates spatial embedding layers with diverse kernel sizes to capture multi-scale local context and model global spatial dependencies (as shown in Fig.~\ref{fig:ms_seffn}). It also employs recursive skip connections and layer normalization to enhance feature discrimination and enable adaptive input-residual fusion for better optimization~\cite{liu2020rethinking}. Given the partitioned feature $CF_i$ in MSFFN, the recursive fusion process is formulated as follows:
\begin{equation}
	{CF_i}^{0}=\operatorname{Dconv}(CF_i),
\end{equation}
\begin{equation}
	{CF_i}^{r+1}=\operatorname{LN}(CF_i+{CF_i}^{r}), r=0,1,2,
\end{equation}
where $\operatorname{Dconv}(\cdot)$ represents the depth-wise convolution, and $\operatorname{LN}(\cdot)$ denotes the layer normalization. To maintain spatial consistency, MSFFN output is first fused with $\hat{E}_i$, then upsampled to original resolution via bilinear interpolation and refined by a convolution, together forming the patch recovery layer before entering PCFF.
Following this, we obtain a smoothed motion field $\tilde{F}^l$, which is further refined by PAR to recover implicitly rectified motion features $M^{l+1}$. Based on the residual information between $M^l$ and $M^{l+1}$, we predict $\hat{z}_{cls}^{L}$ and $\hat{z}_{reg}^{L}$ using two ResBlocks and a linear layer.

\subsection{Loss Function}
We optimize SC-Net using a hybrid loss, which combines the correspondence classification $ \mathcal{L}_{cls}$ and the parametric model regression $\mathcal{L}_{reg}$ ~\cite{yi2018lfgc,zhang2019oanet}:
\begin{equation}
	\mathcal{L}=\sum\nolimits_{l=0}^{L-1}(\mathcal{L}_{cls}(\hat{z}_{cls}^{l},z_{cls})+\lambda\mathcal{L}_{reg}(\hat{E}^l,E)),
\end{equation}
where $L$ is the number of rectifying layers; $\lambda$ is a balance weight; $z_{cls}$ denotes weakly correspondence labels derived by the epipolar distance threshold $10^{-4}$;
$\hat{E}^l$ and $E$ represent the predicted and the ground-truth essential matrix, respectively.
To alleviate label ambiguity, we adopt the adaptive temperature $\tau$ from~\cite{zhao2021clnet} in the classification loss: 
\begin{equation}
	\mathcal{L}_{cls}(\hat{z}_{cls}^{l},z_{cls})=\mathcal{H}(\tau\odot\hat{z}_{cls}^{l},z_{cls}),
\end{equation}
where $\mathcal{H}(\cdot, \cdot)$ is the binary cross-entropy. 
The regression loss follows~\cite{ranftl2018dfe,zhang2019oanet}.

\section{Experiments}
\subsection{Implement Details}
\label{sec:implement}
We establish up to $N=2000$ initial correspondences using SIFT and nearest neighbor matching and additionally use RootSIFT~\cite{arandjelovic2012rootsift} and SuperPoint~\cite{detone2018superpoint} to further validate the generalization ability of the comparison methods.
The model takes normalized input correspondences, scaled to $[-1,1]$ according to the image size or intrinsics. We employ six rectifying layers ($L=6$) with $K=16$ for efficiency and four heads in all GATs. In MFM, input features are split into $P=16$ patches. 
We use the Adam optimizer~\cite{kingma2015adam} with an initial learning rate of $10^{-4}$ , decaying by a factor of $9.6{\times}10^{-5}$ over the first $80k$ steps.
SC-Net is trained for $500k$ steps, with $\lambda$ increasing from $0$ to $0.5$ after $20k$ steps.
All experiments are conducted on Ubuntu 18.04 with an RTX3090.

\subsection{Datasets and Evaluation Protocols}
Following~\cite{zhang2019oanet}, we evaluate SC-Net on both outdoor and indoor scenes. For the outdoor scenes, we use a subset of the YFCC100M dataset~\cite{thomee2016yfcc100m}, which includes 72 image sequences of tourist landmarks, with 68 used as the known scenes and 4 as the unknown scenes. For the indoor scenes, we use the SUN3D dataset~\cite{xiao2013sun3d}, which is split into 239 sequences for the known scenes and 15 for the unknown scenes. The known scenes are divided into training (60\%), validation (20\%), and testing (20\%), while the unknown scenes are used for generalization evaluation.
SC-Net is evaluated on the following two tasks. For the relative pose estimation task, we report the mean Average Precision (mAP)~\cite{zhang2019oanet} of the maximum angular error in rotation and translation under $5^{\circ}$ and $20^{\circ}$ thresholds, along with the area under the cumulative error (AUC)~\cite{zhang2023convmatchaaai} at the same thresholds for comprehensive comparisons. For the outlier removal task, we use \textit{Precision} (\text{P}), \textit{Recall} (\text{R}) and \textit{F-score} (\text{F}) as the metrics.

\begin{table}[!t]
	\centering
\fontsize{7pt}{8.2pt}\selectfont   
\setlength{\tabcolsep}{2.8pt}      
\renewcommand{\arraystretch}{0.95} 
		\begin{tabular}{cc|ccrcccccrcc}
			\toprule
			\multicolumn{2}{c|}{\multirow{3}[6]{*}{Method}} & \multicolumn{5}{c}{YFCC100M (\%)}     &       & \multicolumn{5}{c}{SUN3D (\%)} \\
			\cmidrule{3-13}    \multicolumn{2}{c|}{} & \multicolumn{2}{c}{Known} &       & \multicolumn{2}{c}{UnKnown} &       & \multicolumn{2}{c}{Known} &       & \multicolumn{2}{c}{Unknown} \\
			\cmidrule{3-4}\cmidrule{6-7}\cmidrule{9-10}\cmidrule{12-13}    \multicolumn{2}{c|}{} & $5^{\circ}$    & $20^{\circ}$   &       & $5^{\circ}$    & $20^{\circ}$   &       & $5^{\circ}$    & $20^{\circ}$   &       & $5^{\circ}$    & $20^{\circ}$ \\
			\midrule
			\multicolumn{2}{c|}{RANSAC} & 5.74  & 16.67 &       & 9.05  & 22.71 &       & 4.43  & 15.38 &       & 2.85  & 11.23 \\
			\multicolumn{2}{c|}{LFGC} & 14.51 & 35.82 &       & 23.71 & 50.57 &       & 11.93 & 36.03 &       & 9.73  & 33.09 \\
			\multicolumn{2}{c|}{DFE} & 19.27 & 42.14 &       & 30.55 & 59.15 &       & 14.18 & 39.14 &       & 12.13 & 26.26 \\
			\multicolumn{2}{c|}{ACNe} & 29.63 & 52.71 &       & 34.00 & 62.98 &       & 19.08 & 46.32 &       & 14.27 & 39.29 \\
			\multicolumn{2}{c|}{OANet} & 33.50 & 57.53 &       & 41.33 & 68.79 &       & 22.41 & 49.23 &       & 17.57 & 42.61 \\
			\multicolumn{2}{c|}{T-Net} & 40.86 & 63.81 &       & 46.74 & 73.11 &       & 23.55 & 50.99 &       & 17.69 & 44.03 \\
			\multicolumn{2}{c|}{MSA-Net} & 37.40 & 60.16 &       & 48.45 & 73.23 &       & 18.51 & 45.74 &       & 15.26 & 41.00 \\
			\multicolumn{2}{c|}{MS\textsuperscript{2}DG-Net} & 37.78 & 62.78 &       & 46.98 & 75.13 &       & 22.93 & 50.67 &       & 17.34 & 43.41 \\
			\multicolumn{2}{c|}{NCMNet} & 50.12 & 71.15 &       & 63.85 & 82.44 &       & 25.68 & 52.20 &       & 20.64 & 46.24 \\
			\multicolumn{2}{c|}{U-Match} & 46.03 & 67.60 &       & 60.25 & 79.70 &       & 26.40 & 53.55 &       & 22.38 & 48.62 \\
			\multicolumn{2}{c|}{ConvMatch} & 43.12 & 65.57 &       & 55.45 & 77.49 &       & 27.45 & 54.65 &       & 22.52 & 48.65 \\
			\multicolumn{2}{c|}{ConvMatch\textsuperscript{+}} & 46.26 & 68.45 &       & 57.08 & 79.17 &       & 28.21 & 55.74 &       & 23.08 & 49.12 \\
			\multicolumn{2}{c|}{PT-Net} & 47.41 & 68.90 &       & 57.58 & 79.39 &       & 27.23 & 54.38 &       & 22.88 & 48.52 \\
			\multicolumn{2}{c|}{DeMatch} & 47.53 & 69.08 &       & 59.98 & 79.97 &       & 28.50 & 55.61 &       & 23.51 & 49.84 \\
			\multicolumn{2}{c|}{MSGSA} & 45.94 & 67.67 &       & 57.93 & 78.78 &       & 26.47 & 54.26 &       & 21.01 & 47.43 \\
			\multicolumn{2}{c|}{BCLNet} & \underline{53.21} & \underline{73.48} &       & \underline{67.85} & \underline{84.57} &       & 24.32 & 51.24 &       & 20.06 & 45.83 \\
			\multicolumn{2}{c|}{CGR-Net} & 52.40 & 72.97 &       & 65.97 & 83.62 &       & 26.48 & 53.41 &       & 21.69 & 47.94 \\
			\multicolumn{2}{c|}{DeMo} & 49.78 & 71.33 &       & 63.45 & 82.91 &       & \underline{30.16} & \underline{57.33} &       & \underline{24.00} & \underline{50.44} \\
			\midrule
			\multicolumn{2}{c|}{SC-Net (ours)} & \textbf{64.32} & \textbf{80.40} &       & \textbf{71.75} & \textbf{86.49} &       & \textbf{33.56} & \textbf{59.74} &       & \textbf{25.94} & \textbf{52.05} \\
			\bottomrule
		\end{tabular}
    \caption{Quantitative results of the relative pose estimation task on YFCC100M and SUN3D, presented as mAP@$5^{\circ}$ (\%) and mAP@$20^{\circ}$ (\%). The optimal and suboptimal indicators are shown in \textbf{bold} and \underline{underlined}, respectively.}
    \label{tab:map5}
\end{table}

\subsection{Relative Pose Estimation}
\label{sec:pose}
The relative pose estimation task aims to estimate the positional changes (rotation and translation) between two cameras capturing an image pair. 
We compare SC-Net with previous state-of-the-art MLP-based methods including LFGC~\cite{yi2018lfgc}, DFE~\cite{ranftl2018dfe}, ACNe~\cite{sun2020acne}, OANet~\cite{zhang2019oanet}, T-Net~\cite{zhong2021tnet}, MSA-Net~\cite{zheng2022msanet}, MS\textsuperscript{2}DG-Net~\cite{dai2022ms2dg}, NCMNet~\cite{liu2023ncmnet}, U-Match~\cite{li2023umatch}, MSGSA~\cite{lin2024msgsa}, BCLNet~\cite{miao2024bclnet}, and CGR-Net~\cite{yang2024cgrnet}, and motion-based methods including ConvMatch~\cite{zhang2023convmatchaaai}, ConvMatch$^{+}$~\cite{zhang2023convmatchtpami}, PT-Net~\cite{gong2024ptnet}, DeMatch~\cite{zhang2024dematch} and DeMo~\cite{lu2025demo} on both known and unknown scenes from YFCC100M and SUN3D.
The comparative results are shown in Table \ref{tab:map5}. We observe that the proposed SC-Net achieves exceptional performance across all scenarios. In terms of mAP at $5^\circ$, our method improves performance by $11.11\%$, $3.90\%$, $3.40\%$, and $1.94\%$, respectively, across four different scenes compared to the suboptimal results. 
It is evident that, although motion-based methods do not perform as well as several recent MLP-based methods (e.g., BCLNe and CGR-Net) on YFCC100M, they generally exhibit superior performance on SUN3D, demonstrating the robustness of the motion-based framework.
By the improvements of spatially adaptive focusing and spatial-channel interaction, SC-Net further releases the potential of this framework, achieving more accurate pose estimation. 
AUC results for several representative learning-based methods are reported in Table \ref{tab:auc}. SC-Net consistently outperforms all the competing methods across all evaluation metrics.

\begin{table}[!t]
	\centering
\fontsize{7pt}{8.2pt}\selectfont   
\setlength{\tabcolsep}{2.8pt}      
\renewcommand{\arraystretch}{0.95} 
		\begin{tabular}{cc|ccrcccccrcc}
			\toprule
			\multicolumn{2}{c|}{\multirow{3}[6]{*}{Method}} & \multicolumn{5}{c}{YFCC100M (\%)}     &       & \multicolumn{5}{c}{SUN3D (\%)} \\
			\cmidrule{3-13}    \multicolumn{2}{c|}{} & \multicolumn{2}{c}{Known} &       & \multicolumn{2}{c}{UnKnown} &       & \multicolumn{2}{c}{Known} &       & \multicolumn{2}{c}{Unknown} \\
			\cmidrule{3-4}\cmidrule{6-7}\cmidrule{9-10}\cmidrule{12-13}    \multicolumn{2}{c|}{} & 5°    & 20°   &       & 5°    & 20°   &       & 5°    & 20°   &       & 5°    & 20° \\
			\midrule
			\multicolumn{2}{c|}{OANet} & 14.49 & 48.04 &       & 17.00 & 58.61 &       & 8.04  & 40.14 &       & 6.09  & 34.08 \\
			\multicolumn{2}{c|}{T-Net} & 18.20 & 54.48 &       & 20.70 & 62.94 &       & 9.03  & 42.08 &       & 6.62  & 35.66 \\
			\multicolumn{2}{c|}{MSA-Net} & 15.69 & 49.68 &       & 20.95 & 62.31 &       & 7.24  & 38.54 &       & 6.07  & 34.32 \\
			\multicolumn{2}{c|}{MS\textsuperscript{2}DG-Net} & 15.36 & 52.18 &       & 18.84 & 63.27 &       & 8.51  & 41.81 &       & 6.24  & 35.22 \\
			\multicolumn{2}{c|}{NCMNet} & 25.02 & 61.31 &       & 32.40 & 71.82 &       & 9.96  & 42.88 &       & 7.92  & 38.16 \\
			\multicolumn{2}{c|}{U-Match} & 21.62 & 57.77 &       & 29.57 & 68.95 &       & 10.20 & 44.43 &       & 8.40  & 40.08 \\
			\multicolumn{2}{c|}{ConvMatch} & 19.95 & 55.71 &       & 26.68 & 66.82 &       & 10.84 & 45.43 &       & 8.68  & 40.13 \\
			\multicolumn{2}{c|}{ConvMatch\textsuperscript{+}} & 21.40 & 58.26 &       & 27.31 & 68.18 &       & 11.24 & 46.44 &       & 8.84  & 40.53 \\
			\multicolumn{2}{c|}{PT-Net} & 22.30  & 59.07  &       & 27.62  & 68.52  &       &   10.67   &  45.11  &       &  8.59  &  39.85 \\
			\multicolumn{2}{c|}{DeMatch} & 22.40 & 59.02 &       & 30.00 & 69.35 &       & 11.36 & 46.34 &       & 9.20 & 41.24 \\
			\multicolumn{2}{c|}{MSGSA} & 21.99 & 58.26 &       & 28.23 & 68.31 &       & 10.43 & 44.99 &       & 7.81  & 39.00 \\
			\multicolumn{2}{c|}{BCLNet} & \underline{27.48} & \underline{63.63} &       & \underline{37.22} & \underline{74.44} &       & 9.64  & 42.37 &       & 7.72  & 37.62 \\
			\multicolumn{2}{c|}{CGR-Net} & 26.45 & 62.97 &       & 35.01 & 73.17 &       & 10.60 & 44.29 &       & 8.50  & 39.50 \\
			\multicolumn{2}{c|}{DeMo} & 24.10 & 61.15 &       & 31.65 & 71.97 &       & \underline{12.17} & \underline{47.89} &       & \underline{9.35} & \underline{41.74} \\
			\midrule
			\multicolumn{2}{c|}{SC-Net (ours)} & \textbf{35.27} & \textbf{70.47} &       & \textbf{40.30} & \textbf{76.44} &       & \textbf{14.47} & \textbf{50.34} &       & \textbf{10.36} & \textbf{43.28} \\
			\bottomrule
		\end{tabular}
    \caption{Quantitative results of the relative pose estimation task on YFCC100M and SUN3D, presented as AUC@$5^{\circ}$ (\%) and AUC@$20^{\circ}$ (\%). The optimal and suboptimal indicators are shown in \textbf{bold} and \underline{underlined}, respectively.}
    \label{tab:auc}
\end{table}

\begin{table*}[!t]
	\centering
\fontsize{7pt}{8.2pt}\selectfont   
\setlength{\tabcolsep}{7pt}      
\renewcommand{\arraystretch}{0.95} 
		\begin{tabular}{cc|cccccccccccc}
			\toprule
			\multicolumn{2}{c|}{\multirow{3}[6]{*}{Method}} & \multicolumn{6}{c|}{YFCC100M (\%)}            & \multicolumn{6}{c}{SUN3D (\%)} \\
			\cmidrule{3-14}    \multicolumn{2}{c|}{} & \multicolumn{3}{c|}{Known} & \multicolumn{3}{c|}{Unknown} & \multicolumn{3}{c|}{Known} & \multicolumn{3}{c}{Unknown} \\
			\cmidrule{3-14}    \multicolumn{2}{c|}{} & P (\%) & R (\%) & \multicolumn{1}{c|}{F (\%)} & P (\%) & R (\%) & \multicolumn{1}{c|}{F (\%)} & P (\%) & R (\%) & \multicolumn{1}{c|}{F (\%)} & P (\%) & R (\%) & F (\%) \\
			\midrule
			\multicolumn{2}{c|}{RANSAC~\cite{fischler1981random}} & 47.35 & 52.39 & 49.47 & 43.55 & 50.65 & 46.83 & 51.87 & 56.27 & 53.98 & 44.87 & 48.82 & 46.76 \\
			\multicolumn{2}{c|}{LFGC~\cite{yi2018lfgc}} & 54.43 & 86.88 & 66.93 & 52.84 & 85.68 & 65.37 & 53.70 & 87.03 & 66.42 & 46.11 & 83.92 & 59.52 \\
			\multicolumn{2}{c|}{DFE~\cite{ranftl2018dfe}} & 56.72 & 87.16 & 68.72 & 54.00 & 85.56 & 66.21 & 53.96 & 87.23 & 66.68 & 46.18 & 84.01 & 59.60 \\
			\multicolumn{2}{c|}{ACNe~\cite{sun2020acne}} & 60.02 & 88.99 & 71.69 & 55.62 & 85.47 & 67.39 & 54.11 & 88.46 & 67.15 & 46.16 & 84.01 & 59.58 \\
			\multicolumn{2}{c|}{OANet~\cite{zhang2019oanet}} & 61.14 & 88.16 & 69.73 & 57.90 & 85.07 & 66.53 & 54.43 & 88.08 & 63.72 & 46.50 & 83.83 & 56.32 \\
			\multicolumn{2}{c|}{T-Net~\cite{zhong2021tnet}} & 61.18 & 89.94 & 70.47 & 57.18 & 87.01 & 66.73 & 55.01 & 88.36 & 64.18 & 46.50 & 83.98 & 56.33 \\
			\multicolumn{2}{c|}{MSA-Net~\cite{zheng2022msanet}} & 59.27 & 90.28 & 68.92 & 56.49 & 88.60 & 66.46 & 56.09 & 87.57 & 64.71 & 48.64 & 83.81 & 57.89 \\
			\multicolumn{2}{c|}{MS\textsuperscript{2}DG-Net~\cite{dai2022ms2dg}} & 64.24 & 89.31 & 72.49 & 60.38 & 86.71 & 68.96 & 55.58 & 89.01 & 64.63 & 47.42 & 84.50 & 57.12 \\
			\multicolumn{2}{c|}{U-Match~\cite{li2023umatch}} & 63.29 & 92.09 & 72.56 & 61.02 & 90.67 & 70.61 & 55.29 & \underline{89.35} & 64.53 & 47.69 & \underline{85.60} & 57.53 \\
			\multicolumn{2}{c|}{ConvMatch~\cite{zhang2023convmatchaaai}} & 63.15 & 91.19 & 72.21 & 60.22 & 89.48 & 69.66 & 55.79 & 89.23 & 64.89 & 48.13 & 85.54 & 57.87 \\
			\multicolumn{2}{c|}{ConvMatch\textsuperscript{+}~\cite{zhang2023convmatchtpami}} & 63.24 & 91.90 & 72.52 & 60.10 & 89.35 & 69.58 & 55.29 & 89.33 & 64.56 & 47.22 & 85.42 & 57.15 \\
			\multicolumn{2}{c|}{MSGSA~\cite{lin2024msgsa}} & 63.48 & 91.04 & 74.80 & 60.43 & 89.01 & 71.98 & 55.92 & 88.56 & 68.55 & 47.99 & 84.32 & 61.22 \\
			\multicolumn{2}{c|}{PT-Net~\cite{gong2024ptnet}} & 63.89 & 91.57 & 72.99 & 60.45 & 89.29 & 69.88 & 55.41 & 89.17 & 64.66 & 47.45 & 85.52 & 57.39 \\
			\multicolumn{2}{c|}{DeMatch~\cite{zhang2024dematch}} & 61.32 & \textbf{92.76} & 71.18 & 58.74 & \underline{91.04} & 68.91 & 56.00 & 88.90 & 65.10 & 48.27 & 85.24 & 58.08 \\
			\multicolumn{2}{c|}{NCMNet~\cite{liu2023ncmnet}} & 77.69 & 81.27 & 79.05 & 76.58 & 78.58 & 77.33 & 66.23 & 74.69 & 69.46 & 61.19 & 69.07 & \underline{64.25} \\
			\multicolumn{2}{c|}{BCLNet~\cite{miao2024bclnet}} & \underline{78.36} & 82.23 & \underline{79.87} & \underline{77.90} & 80.07 & \underline{78.73} & 66.20 & 74.12 & 69.19 & 61.14 & 68.33 & 63.92 \\
			\multicolumn{2}{c|}{CGR-Net~\cite{yang2024cgrnet}} & 77.88 & 81.41 & 79.22 & 77.06 & 79.11 & 77.84 & \underline{66.46} & 74.46 & \underline{69.51} & \underline{61.24} & 68.85 & 64.18 \\
			\multicolumn{2}{c|}{DeMo~\cite{lu2025demo}} & 64.42 & \underline{92.68} & 73.71 & 61.41 & \textbf{91.28} & 71.22 & 56.08 & \textbf{89.66} & 65.28 & 47.87 & \textbf{85.74} & 57.75 \\			
			\midrule
			\multicolumn{2}{c|}{SC-Net (ours)} & \textbf{83.91} & 82.96 & \textbf{82.82} & \textbf{82.14} & 79.05 & \textbf{80.06} & \textbf{73.13} & 74.01 & \textbf{72.49} & \textbf{68.49} & 69.88 & \textbf{68.25} \\
			\bottomrule
		\end{tabular}
	\caption{Quantitative results of the outlier removal task on YFCC100M and SUN3D, presented as \textit{Precision} (\%), \textit{Recall} (\%) and \textit{F-score} (\%). The optimal and suboptimal indicators are shown in \textbf{bold} and \underline{underlined}, respectively.}
	\label{tab:prf}
\end{table*}
\begin{figure}[!t]
	\centering
	\begin{subfigure}[t]{0.23\textwidth}
        \centering
        \includegraphics[width=\linewidth]{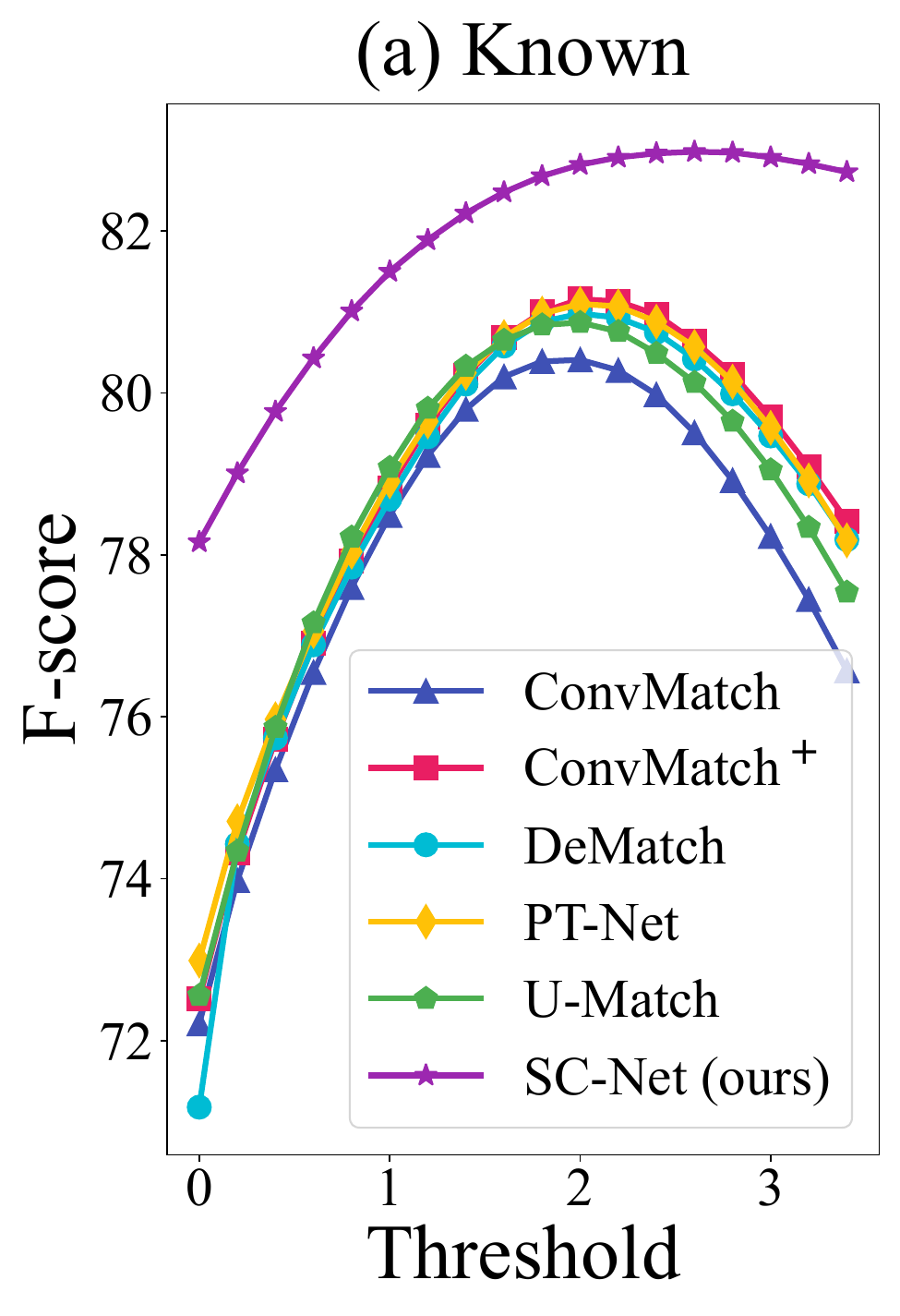}
    \end{subfigure}
	\hfill
    \begin{subfigure}[t]{0.23\textwidth}
        \centering
        \includegraphics[width=\linewidth]{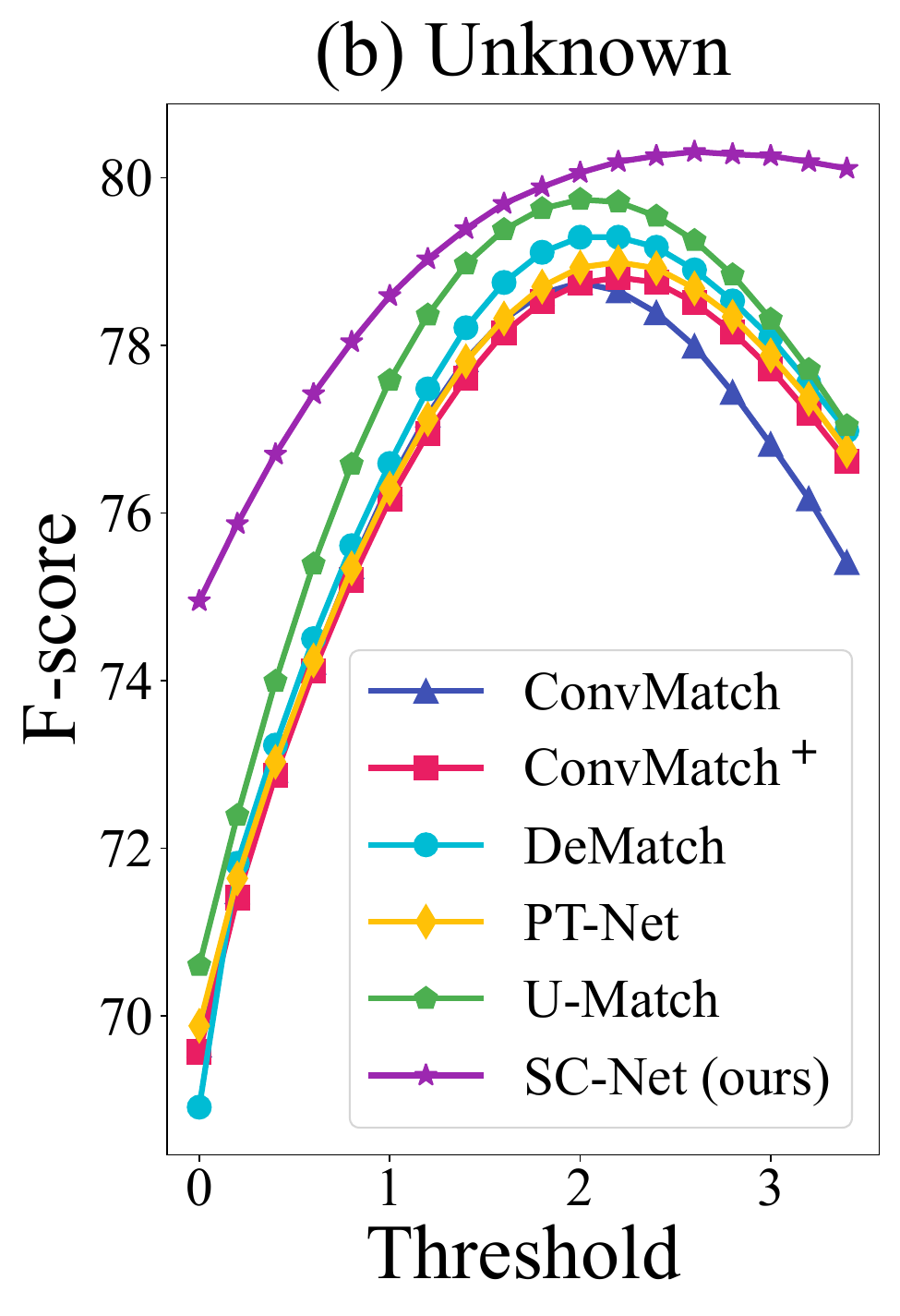}
    \end{subfigure}
	\caption{Comparison of F-score results for different models under the different logit thresholds on the known (a) and unknown (b) scenes in YFCC100M.}
	\label{fig:fscores}
\end{figure}
\begin{figure}[!t]
	\centering
	\includegraphics[width=0.48\textwidth]{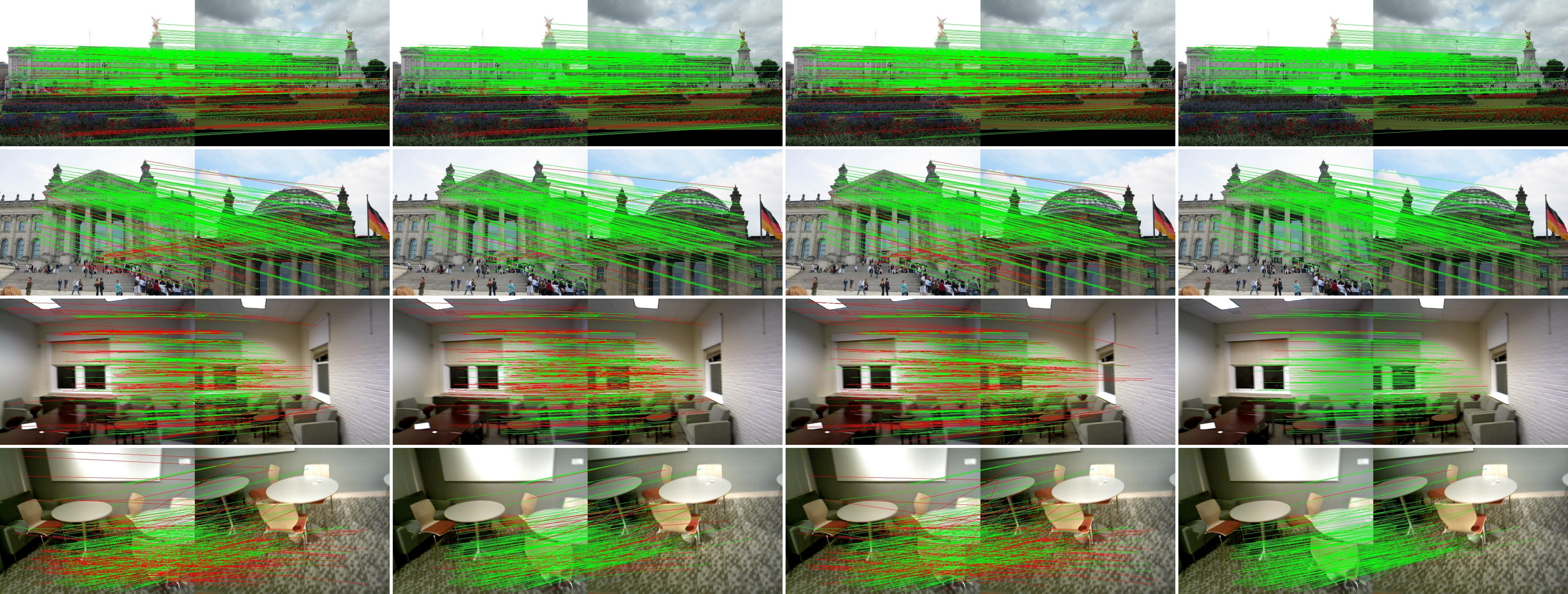}
	\begin{tabular}{cccc}
		ConvMatch & ~~~~\footnotesize MSGSA & \hspace{6pt} ~~~~\footnotesize PT-Net & \hspace{6pt} ~~\footnotesize SC-Net (Ours)
	\end{tabular}
    \caption{Qualitative results of outlier removal (1\textsuperscript{st} and 2\textsuperscript{nd} rows show the outdoor scenes from YFCC100M, while 3\textsuperscript{rd} and 4\textsuperscript{th} rows show the indoor scenes from SUN3D). False matches are marked in red and correct matches in green.}
	\label{fig:match_vis}
\end{figure}

\subsection{Outlier Removal}
The outlier removal task aims to identify and eliminate incorrect correspondences between image pairs, improving the accuracy of subsequent processing steps.
As shown in Table \ref{tab:prf}, these methods (i.e., NCMNet~\cite{liu2023ncmnet}, BCLNet~\cite{miao2024bclnet}, and CGR-Net~\cite{yang2024cgrnet}) perform correspondence classification based on the epipolar distance of each correspondence with respect to the estimated parametric model, whereas other methods (i.e., LFGC~\cite{yi2018lfgc}, DFE~\cite{ranftl2018dfe}, ACNe~\cite{sun2020acne}, OANet~\cite{zhang2019oanet}, T-Net~\cite{zhong2021tnet}, MSA-Net~\cite{zheng2022msanet}, MS\textsuperscript{2}DG-Net~\cite{dai2022ms2dg}, U-Match~\cite{li2023umatch}, MSGSA~\cite{lin2024msgsa}, DeMatch~\cite{zhang2024dematch}, ConvMatch~\cite{zhang2023convmatchaaai}, ConvMatch$^{+}$~\cite{zhang2023convmatchtpami}, and PT-Net~\cite{gong2024ptnet}) identify inliers based on the predicted logits of correspondences. 
It is obvious that the logit-based methods generally achieve higher recall but relatively lower precision compared to distance-based ones. As a result, they often fail to achieve F-scores comparable to those of distance-based methods. However, we empirically observe that the inlier threshold of logits significantly affects the classification performance. Therefore, we test different thresholds for validating the effectiveness of our method. The visualization results of the F-score under different thresholds are presented in Fig.~\ref{fig:fscores}. Our method outperforms other recent logit-based methods across all threshold settings and achieves the highest peak F-score, even significantly surpassing the distance-based ones at certain thresholds. 
Compared to the suboptimal results of other alternatives, SC-Net obtains an improvement of $1.33\%$ to $4.00\%$ in the F-score metric. We further present a visualization of several matching results in Fig.~\ref{fig:match_vis}. Our method effectively combines strong mismatch filtering capabilities with the retention of a substantial number of correct matches, leading to more precise pose estimation as in the last section.

\section{Discussions}
\subsection{Generalization Ability}
\label{sec:generalization}
We combine other keypoint detectors~\cite{arandjelovic2012rootsift,detone2018superpoint} to evaluate the generalization ability of our method. The results are shown in Table \ref{tab:generalization}. All models are trained with SIFT correspondences but tested with correspondences extracted by other detectors. We can observe that our method outperforms other alternatives in all cases, which effectively demonstrates its robustness and generalization ability.
We further evaluate SC-Net on the image matching benchmark PhotoTourism~\cite{jin2021image}, with results summarized in Table~\ref{tab:IMC-Pt}. Table~\ref{tab:IMC-Pt} reports cross-dataset generalization results on PhotoTourism and SUN3D using both SIFT and SuperPoint matches. SC-Net achieves the best performance under both descriptors, surpassing previous methods such as BCLNet, ConvMatch\textsuperscript{+} and DeMo. In particular, its strong performance on SuperPoint—characterized by dense and irregular keypoints—demonstrates SC-Net’s superior spatial awareness and generalization. In terms of computational cost, SC-Net maintains moderate FLOPs and peak memory usage, while being significantly more efficient than heavier models like BCLNet and CGR-Net. Overall, SC-Net offers an effective trade-off between accuracy and efficiency.

\subsection{Parameter Setting}
We conduct comparative experiments on SC-Net with varying parameters, including different numbers of rectifying layers and grids. As shown in Table \ref{tab:parameters}, the optimal combination of $L=6$ and $K=16$ yields the best performance. 
Increasing the number of rectifying layers initially improves pose estimation accuracy, but beyond a certain point, performance deteriorates because excessive layers tend to filter out motions with slight deviations from the ideal, reducing reliable correspondences.
A larger $K$ value helps refine the motion field but may also introduce additional noise during regularization, thereby complicating subsequent rectification.

\begin{table}[!t]
	\centering
\fontsize{7pt}{8.2pt}\selectfont   
\setlength{\tabcolsep}{3pt}      
\renewcommand{\arraystretch}{0.95} 
			\begin{tabular}{cc|ccccc}
				\toprule
				\multicolumn{2}{c|}{\multirow{2}[4]{*}{Method}} & \multicolumn{2}{c}{YFCC100M (\%)} &  & \multicolumn{2}{c}{SUN3D (\%)} \\
				\cmidrule{3-4}\cmidrule{6-7}    \multicolumn{2}{c|}{} & RootSIFT & SP & & RootSIFT & SP \\
				\midrule
				\multicolumn{2}{c|}{OANet~\cite{zhang2019oanet}} & 41.73 & 19.12 & & 17.41 & 6.60 \\
				\multicolumn{2}{c|}{T-Net~\cite{zhong2021tnet}} & 49.35 & 23.15 & & 17.93 & 7.47 \\
				\multicolumn{2}{c|}{MSA-Net~\cite{zheng2022msanet}} & 48.45 & 19.78 & & 16.85 & 6.40 \\
				\multicolumn{2}{c|}{MS\textsuperscript{2}DG-Net~\cite{dai2022ms2dg}} & 45.95 & 19.12 & & 17.50 & 7.08 \\
				\multicolumn{2}{c|}{NCMNet~\cite{liu2023ncmnet}} & 63.38 & 23.55 & & 21.44 & 9.57 \\
				\multicolumn{2}{c|}{U-Match~\cite{li2023umatch}} & 60.85 & 26.58 & & 22.47 & 8.46 \\
				\multicolumn{2}{c|}{ConvMatch~\cite{zhang2023convmatchaaai}} & 56.20 & 28.15 & & 22.77 & 12.95 \\
				\multicolumn{2}{c|}{ConvMatch\textsuperscript{+}~\cite{zhang2023convmatchtpami}} & 58.60 & \underline{30.85} & & 23.54 & 13.77 \\
				\multicolumn{2}{c|}{PT-Net~\cite{gong2024ptnet}} & 58.25 & 30.60 & & 22.96 & 13.28 \\
				\multicolumn{2}{c|}{DeMatch~\cite{zhang2024dematch}} & 61.20 & 27.15 & & 24.09 & \underline{14.22} \\
				\multicolumn{2}{c|}{MSGSA~\cite{lin2024msgsa}} & 54.42 & 26.35 & & 21.42 & 9.14 \\
				\multicolumn{2}{c|}{BCLNet~\cite{miao2024bclnet}} & \underline{68.37} & 24.37 & & 20.66 & 8.17 \\
				\multicolumn{2}{c|}{CGR-Net~\cite{yang2024cgrnet}} & 67.53 & 21.30 & & 22.42 & 8.99 \\
				\multicolumn{2}{c|}{DeMo~\cite{lu2025demo}} & 63.98 & 20.85 & & \underline{24.51} & 9.92 \\
				\midrule
				\multicolumn{2}{c|}{SC-Net (ours)} & \textbf{71.97} & \textbf{44.08} & & \textbf{26.34} & \textbf{23.52} \\
				\bottomrule
			\end{tabular}
	\caption{Intra-dataset generalization evaluation. All models are trained with SIFT matches and then tested with RootSIFT and SuperPoint (SP). mAP@$5^{\circ}$ is reported.}
	\label{tab:generalization}
\end{table}

\subsection{Ablation Studies}
In this section, we conduct ablation studies on the outdoor scenes from YFCC100M to verify the effectiveness of each component. 
Notably, by modifying ConvMatch to use soft-mask Softmax for confidence calculation in the weighted 8-point algorithm, the baseline achieves performance gains consistent with the findings of MSGSA.
As a foundational component of BFA, the hierarchical encoder-decoder (HED) performs multi-level encoding and decoding of the input motion field, producing more hierarchical and enriched feature representations. This enables it to capture subtle variations and correct the motion field based on high-level motion patterns. Building upon the HED foundation, the motion feature modulator (MFM) dynamically adjusts motion features based on channel-wise feature distributions, effectively reducing ambiguity. In AFR, the soft filtering mechanism (SF) incorporates the classification prediction from the previous layer to suppress noisy motion samples, enhancing the robustness of our network. Meanwhile, position-aware attention (PA) improves spatial awareness by incorporating spatial correlations into the attention mechanism, enabling more precise motion field initialization. 
As shown in Table \ref{tab:ablation}, the performance of our network progressively improves with the incremental addition of each component. Ultimately, the integration of all components leads to the best overall results.

\begin{table}[!t]
	\centering
\fontsize{7pt}{8.2pt}\selectfont   
\setlength{\tabcolsep}{2.5pt}      
\renewcommand{\arraystretch}{0.95} 
			\begin{tabular}{cc|ccccccccc}
				\toprule
				\multicolumn{2}{c|}{\multirow{2}[4]{*}{Method}} & \multicolumn{2}{c}{PhotoTourism (\%)} &       & \multicolumn{2}{c}{SUN3D (\%)} & & \multicolumn{1}{c}{\multirow{2}{*}{\makecell{FLOPs\\(G)}}} & \multicolumn{1}{c}{\multirow{2}{*}{\makecell{Time\\(ms)}}} & \multicolumn{1}{c}{\multirow{2}{*}{\makecell{Memory\\(GB)}}} \\
				\cmidrule{3-4}\cmidrule{6-7}    \multicolumn{2}{c|}{} & SIFT  & SP    &       & SIFT  & SP  &  &       &       &  \\
				\midrule
				\multicolumn{2}{c|}{OANet} & 33.43 & 19.73 &       & 3.89  & 3.62 &  & 117.80 & \textbf{30.29} & \textbf{0.33} \\
				\multicolumn{2}{c|}{T-Net} & 42.27 & 28.24 &       & 4.12  & 3.75 & & 173.10 & 56.36 & 0.43 \\
				\multicolumn{2}{c|}{MSA-Net} & 35.98 & 19.83 &       & 2.44  & 3.65  & & \textbf{100.22} & 52.16 & 0.43 \\
				\multicolumn{2}{c|}{MS\textsuperscript{2}DG-Net} & 35.72 & 24.07 &       & 3.71  & 3.30  & & 322.81 & 163.90 & 2.52 \\
				\multicolumn{2}{c|}{NCMNet} & 54.11 & 34.97 &       & 6.05  & 4.09  & & 557.92 & 434.71 & 2.46 \\
				\multicolumn{2}{c|}{U-Match} & 45.48 & 30.62 &       & 6.77  & 3.61  & & 239.40 & 82.49 & 0.61 \\
				\multicolumn{2}{c|}{ConvMatch} & 45.66 & 33.86 &       & 6.33  & 5.38  & & 242.14 & 64.20 & 0.65 \\
				\multicolumn{2}{c|}{ConvMatch\textsuperscript{+}} & 48.23 & \underline{39.22} &       & 6.95  & \underline{6.24} & & 344.27 & 73.11 & 0.68 \\
				\multicolumn{2}{c|}{PT-Net} & 49.81 & 39.19 &       & 5.49  & 6.00  & & 193.25 & 67.04 & 0.66 \\
				\multicolumn{2}{c|}{DeMatch} & 49.69 & 35.23 &       & 7.60  & 4.88  & & 150.18 & 37.24 & 0.46 \\
				\multicolumn{2}{c|}{MSGSA} & 48.37 & 33.74 &       & 5.98  & 4.20  & & 299.43 & 107.65 & 0.92 \\
				\multicolumn{2}{c|}{BCLNet} & \underline{56.58} & 32.11 &       & 5.05  & 2.95  & & 644.05 & 290.87 & 1.96 \\
				\multicolumn{2}{c|}{CGR-Net} & 55.91 & 30.10 &       & 5.85  & 3.68  & & 424.40 & 217.50 & 1.98 \\
				\multicolumn{2}{c|}{DeMo} & 51.15 & 33.81 &       & \underline{7.65} & 3.83  & & 231.17 & 113.80 & 0.52 \\
				\midrule
				\multicolumn{2}{c|}{SC-Net (ours)} & \textbf{67.58} & \textbf{58.83} &       & \textbf{7.77} & \textbf{7.15} & & 293.84 & 128.55 & 1.13 \\
				\bottomrule
			\end{tabular}%
	\caption{Cross-dataset generalization evaluation. All models are trained on YFCC100M with SIFT matches and then applied in inference to other datasets. mAP@$5^{\circ}$ is reported.}
	\label{tab:IMC-Pt}%
\end{table}%

\begin{table}[!t]
	\centering
\fontsize{7pt}{8.2pt}\selectfont   
\setlength{\tabcolsep}{6pt}      
\renewcommand{\arraystretch}{0.93} 
		\begin{tabular}{c|cc}
			\toprule
			Cases & mAP@$5^\circ$ & mAP@$20^\circ$ \\
			\midrule
			$L=4$   & 60.79 & \underline{78.33} \\
			\boldmath{$L=6$}   &\textbf{64.35} & \textbf{80.38} \\
			$L=8$   & \underline{63.96} & \textbf{80.38} \\
			\midrule
			$K=12$  & 63.24 & 79.84 \\
			\boldmath{$K=16$}  & \textbf{64.35} & \textbf{80.38} \\
			$K=20$  & \underline{63.40} & \underline{80.04} \\
			\bottomrule
		\end{tabular}
	\caption{Comparison results of the relative pose estimation task for the different number of layers $L$ and the number of grids $K$ in YFCC100M.}
	\label{tab:parameters}
\end{table}
\begin{table}[!t]
	\centering
\fontsize{7pt}{8.2pt}\selectfont   
\setlength{\tabcolsep}{5pt}      
\renewcommand{\arraystretch}{0.93} 
			\begin{tabular}{ccccc|cc}
				\cmidrule{1-7}    Baseline & HED   & MFM   & SF    & PA    & mAP@$5^\circ$ & mAP@$20^\circ$ \\
				\midrule
				\checkmark     &       &       &       &       & 46.09 & 67.59 \\
				\checkmark     & \checkmark     &       &       &       & 57.96 & 76.72 \\
				\checkmark     & \checkmark     & \checkmark     &       &       & 59.60 & 77.78 \\
				\checkmark     & \checkmark     & \checkmark     & \checkmark     &       & \underline{61.96} & \underline{78.89} \\
				\checkmark     & \checkmark     & \checkmark     & \checkmark     & \checkmark     & \textbf{64.35} & \textbf{80.38} \\
				\bottomrule
			\end{tabular}
	\caption{Ablation results in YFCC100M. BFA consists of \textbf{HED} and \textbf{MFM}. AFR includes \textbf{SF} and \textbf{PA}.}
	\label{tab:ablation}
\end{table}

\section{Conclusion}
In this paper, we propose SC-Net, an effective network designed for challenging correspondence learning tasks. To address key limitations in the existing framework, we design AFR to generate cleaner initial motion fields and construct BFA to capture long-range dependencies across spatial and channel dimensions. These components collectively enhance SC-Net’s ability to refine motion fields. We validate SC-Net through extensive experiments on public benchmarks, which demonstrate superior accuracy, robustness, and generalization over existing state-of-the-art methods. Future work includes improving SC-Net’s efficiency and extending it to low-overlap or dynamic scenarios.

\section{Acknowledgments}
This work was supported in part by National Natural Science Foundation of China (Nos. U22A2095, 62476112, 62332007, U22B2028, 62501343, 62172197); in part by Guangdong Basic and Applied Basic Research Foundation (Nos. 2024A1515011740, 2025A1515010181); 
in part by Natural Science Foundation of Shandong Province (No. ZR2024QF294); 
in part by Fundamental Research Funds for the Central Universities (Nos. 21624404, 23JNSYS01); 
in part by Science and Technology Major Project of Tibetan Autonomous Region of China (No. XZ202201ZD0006G), National Joint Engineering Research Center of Network Security Detection and Protection Technology, Guangdong Key Laboratory of Data Security and Privacy Preserving, Guangdong Hong Kong Joint Laboratory for Data Security and Privacy Protection, and Engineering Research Center of Trustworthy AI, Ministry of Education.

\bibliography{aaai2026}

\end{document}